\documentclass[acmtog]{acmart}
\acmSubmissionID{TOG-19-0029}

\usepackage{booktabs} % For formal tables
\usepackage{subfig}
\usepackage{float}
\usepackage{color,soul}

\usepackage[ruled]{algorithm2e} % For algorithms

\SetAlFnt{\small}
\SetAlCapFnt{\small}
\SetAlCapNameFnt{\small}
\SetAlCapHSkip{0pt}

\acmJournal{TOG}
\setcopyright{acmcopyright}
\hypersetup{draft}
\begin{document}
% Title portion
\title{Deep Iterative Frame Interpolation for Full-frame Video Stabilization}

\begin{teaserfigure}
	\includegraphics[width=1\linewidth,keepaspectratio]{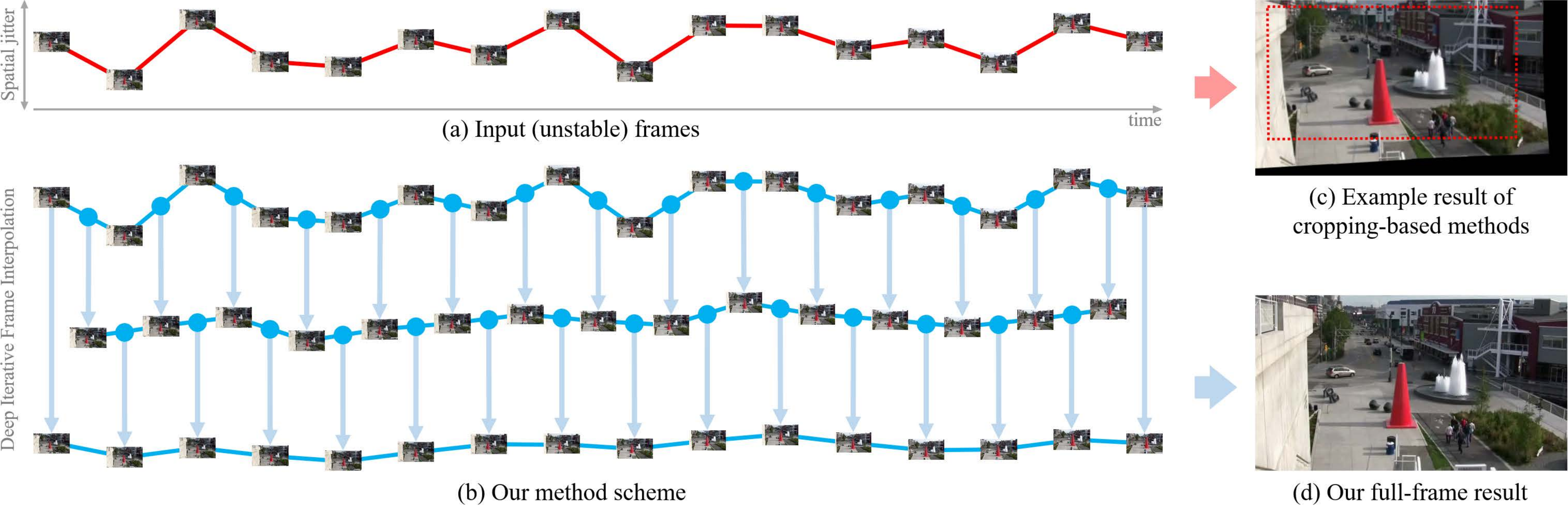}
	\vspace{-7mm}
	\caption{Given (a) an unstable input video, (b) our method utilizes frame interpolation as a means of stabilization. 
	Previous methods mostly involve (c) spatially adjusted frames followed by frame cropping (shown in red dotted lines), whereas (d) our results are full-frame.
	Our method essentially interpolates between two frames with spatial jitter (y-axis, expressed as 1D displacement) across successive frames (x-axis).
	Thus, our frame interpolation method stabilizes frames while interpolating frame boundaries as an added essential effect.
	This diagram illustrates how frame interpolation can lead to stabilization. 
	For our detailed interpolation method, refer to Sec.~\ref{method}
	}
	\label{teaser}
\end{teaserfigure}

% DO NOT ENTER AUTHOR INFORMATION FOR ANONYMOUS TECHNICAL PAPER SUBMISSIONS TO SIGGRAPH 2019!
\author{Jinsoo Choi}
\author{In So Kweon}
%\orcid{1234-5678-9012-3456}
\affiliation{%
  \institution{KAIST}
  \streetaddress{291 Daehak-ro, Yuseong-gu}
  \city{Daejeon}
  %\state{VA}
  \postcode{34141}
  \country{Republic of Korea}
 }
\email{jinsc37@gmail.com}
\email{iskweon77@kaist.ac.kr}
%\author{In So Kweon}
%\affiliation{%
%  \institution{KAIST}
%  %\city{Rocquencourt}
%  \country{Republic of Korea}
%}
%\email{beranger@inria.fr}
%\affiliation{%
% \institution{Rajiv Gandhi University}
%\email{aprna_patel@rguhs.ac.in}
%\affiliation{%
%  \institution{Tsinghua University}
%  \streetaddress{30 Shuangqing Rd}
%  \country{China}
%}
%\email{chan0345@tsinghua.edu.cn}
%\author{Ting Yan}
%\affiliation{%
%  \institution{Eaton Innovation Center}
%  \city{Prague}
%  \country{Czech Republic}}
%\email{yanting02@gmail.com}
%\author{Tian He}
%\affiliation{%
%  \institution{University of Virginia}
%  \department{School of Engineering}
%  \city{Charlottesville}
%  \state{VA}
%  \postcode{22903}
%  \country{USA}
%}
%\affiliation{%
%  \institution{University of Minnesota}
%\email{tinghe@uva.edu}
%\author{Chengdu Huang}
%\author{John A. Stankovic}
%\author{Tarek F. Abdelzaher}
%\affiliation{%
%  \institution{University of Virginia}
%  \department{School of Engineering}
%  \city{Charlottesville}
%  \state{VA}
%  \postcode{22903}
%  \country{USA}
%}

%\renewcommand\shortauthors{Zhou, G. et al}

\begin{abstract}
Video stabilization is a fundamental and important technique for higher quality videos.
Prior works have extensively explored video stabilization, but most of them involve cropping of the frame boundaries and introduce moderate levels of distortion.
We present a novel deep approach to video stabilization which can generate video frames without cropping and low distortion.
The proposed framework utilizes frame interpolation techniques to generate in between frames, leading to reduced inter-frame jitter.
Once applied in an iterative fashion, the stabilization effect becomes stronger.
A major advantage is that our framework is end-to-end trainable in an unsupervised manner.
In addition, our method is able to run in near real-time (15 fps).
To the best of our knowledge, this is the first work to propose an unsupervised deep approach to full-frame video stabilization.
We show the advantages of our method through quantitative and qualitative evaluations comparing to the state-of-the-art methods.
\end{abstract}





\maketitle

\section{Introduction}
%1) X is an important problem 
%2) The core challenges are this and that. 
%3) Previous work on X has addressed these with Y, but the problems with this are Z. 
%4) In this work we do W. 
%5) This has the following appealing properties and our experiments show this and that. 
%You can play with this structure a bit but these core points should be clearly made.

% This is an important problem
As ever growing amounts of video are uploaded to the web, high quality videos are becoming increasingly important and subject to high demands.
Visual stability is an essential aspect of high quality videos especially for hand held and shaky footages.
Thus, video stabilization is a desirable technique for commercial platforms like YouTube, and editing software such as Adobe Premiere.

% Challenges and previous methods
In the past years, we have seen exceptional progress in the field of video stabilization.
Various approaches have been explored including 3D~\cite{liu2009content,liu2012video,zhou2013plane}, 2.5D~\cite{liu2011subspace,liu2013joint,goldstein2012video}, 2D~\cite{liu2014steadyflow,liu2016meshflow,liu2013bundled}, and deep learning based~\cite{wang2018deep,xu2018deep} approaches.
These approaches have shown significant success and continue to affect commercial stabilization algorithms.
Most existing video stabilization methods stabilize videos offline as post-processing.
Although offline methods generally show better stabilization results compared to online methods, recent deep learning-based~\cite{wang2018deep,xu2018deep} approaches have shown promising quality.
The deep learning-based methods take a supervised approach which requires unstable (shaky) and stable (motion smoothed) video paired data.
Thus, they utilize a dataset~\cite{wang2018deep} that has been collected by an unstable and stable cameras capturing the same scene simultaneously.
Moreover, most of the video stabilization approaches including the deep learning methods need to crop the frame boundaries due to temporally \emph{missing view}.
Camera shake causes temporary missing content at the frame boundaries when contrasted to adjacent frames (Fig.~\ref{teaser}~(c)).
Thus, the final step of most state-of-the-art methods involve cropping the frame boundaries.
Cropping causes loss of original content and an inevitable \emph{zoom-in effect}, which is why one of the main goals of video stabilization research is to reduce excessive cropping.

\begin{figure}
	\includegraphics[width=1\linewidth,keepaspectratio]{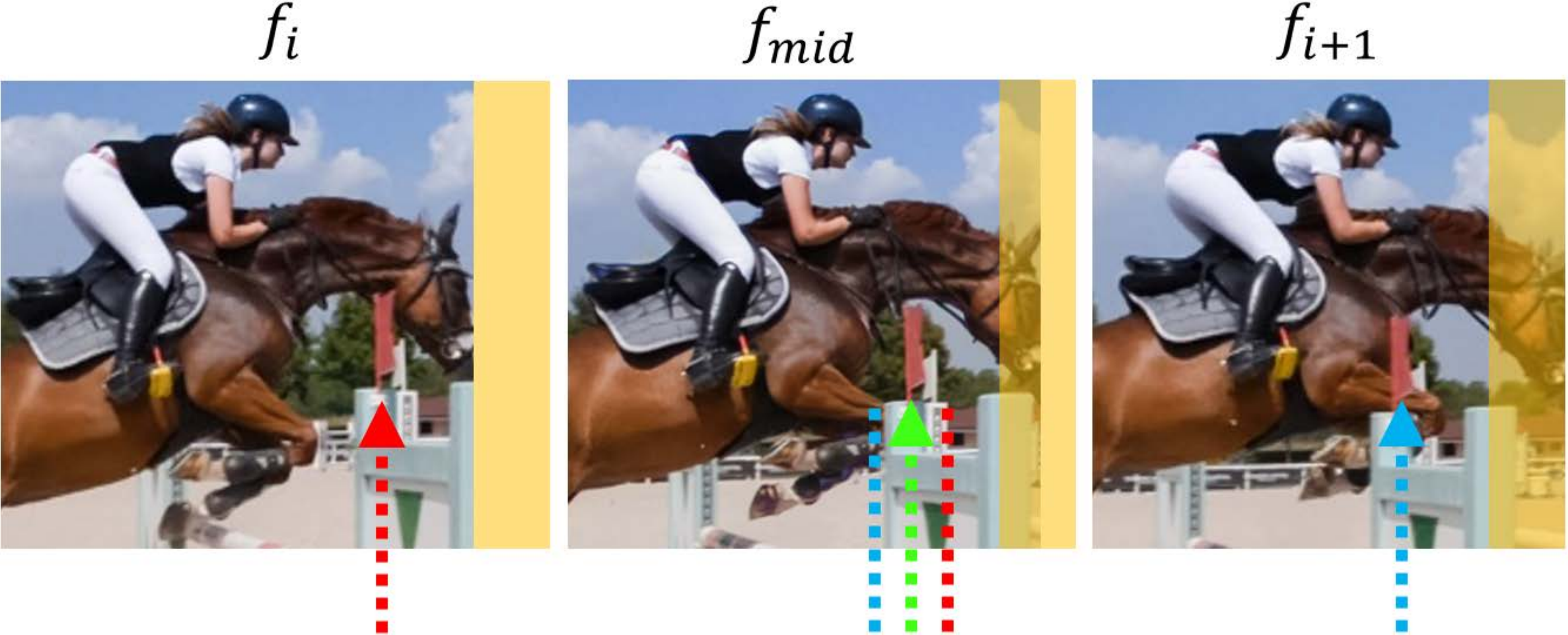}
	%\vspace{-5mm}
	\caption{Our actual frame interpolation ($f_{mid}$) between frames $f_i$ and $f_{i+1}$.
	Ideally, every pixel is interpolated between inter-frame motion.
	For example, the red flag is interpolated halfway (green arrow) between its inter-frame positions (red, blue arrows).
	Also, frame boundaries can be interpolated between unseen (in $f_i$) and seen (in $f_{i+1}$) regions (denoted in yellow) via inter-frame camera motion.}
	\label{interpolation}
	\vspace{-5mm}
\end{figure}

% What we propose
We propose a deep framework which can be trained in an \emph{unsupervised} manner, and thus does not require corresponding stabilized ground truth data.
Our method utilizes frame interpolation to stabilize frames, which in effect \emph{eliminates} cropping.
Essentially, our deep framework learns to generate the ``in between'' middle frame of two sequential frames.
Viewing from the perspective of interpolation, the synthesized middle (i.e. interpolated) frame represents the frame that \emph{would have} been captured in between two sequential frames.
Thus, the interpolated frame represents the exact temporally middle frame, assumed to be \emph{caught} exactly halfway of inter-frame motion \cite{niklaus2017adaptive,niklaus2018context}.
Consequently, sequential generation of middle frames leads to reduced spatial jitter between adjacent frames as shown in Fig.~\ref{teaser}~(b).
Intuitively, frame interpolation can be thought of as linear interpolation (low-pass filter) in the time domain for spatial data sequences. 
When linear interpolation is applied multiple times, the stabilization effect becomes stronger.
For spatial data sequences (i.e frame sequences), interpolation estimates the exact halfway point for essentially every pixel and generates the middle frame as illustrated in Fig.~\ref{interpolation}.
In addition, an important merit of middle frame synthesis is that frame boundaries are synthesized between any inter-frame camera motion, thus filling in temporally missing view and leading to full-frame capability.
Moreover, due to the deep architecture, fast feed forwarding enables our method to run in near real-time (15 fps).

% Contributions, and experiments
Our proposed method, which we call the Deep Iterative FRame INTerpolation (DIFRINT) method, has the following contributions.

1. Our approach enables full-frame (without cropping) video stabilization in near real-time.

2. Our framework is trained in an unsupervised manner, which does not require stable ground truth video counterparts.

3. Our method incorporates frame interpolation techniques, leading to low visual distortion (wobbling artifacts).

To the best of our knowledge, this is the first work to propose a deep unsupervised approach to full-frame video stabilization.
We evaluate our method via quantitative and qualitative comparisons against numerous state-of-the-art methods.
Furthermore, we provide visual comparisons, ablation study analysis, and user preference studies against commercial stabilization methods.

%%%%%%%%%%%%%%%%%%%%%%%%%%%%%%%%%%%%%%%%%%%%%%%%%%%%%%%%%%%%%%%%%%%%%
\section{Related Work}
Video stabilization methods can be categorized into 3D, 2.5D, 2D, and deep learning-based approaches.

3D methods model the camera trajectory in 3D space for stabilization.
To stabilize camera motion, various techniques including Structure from Motion (SfM)~\cite{liu2009content}, depth information~\cite{liu2012video}, 3D plane constraints~\cite{zhou2013plane}, projective 3D reconstruction~\cite{buehler2001non}, light field~\cite{smith2009light}, 3D rotation estimation via gyroscope~\cite{karpenko2011digital,bell2014non,ovren2015gyroscope} has been used.
2.5D approaches use partial information from 3D models, and can handle reconstruction failure cases.
Liu et al.~\shortcite{liu2011subspace} smooths the long feature tracks via subspace constraints to maintain 3D constraints, while its extension~\cite{liu2013joint} applies the approach to stereoscopic videos.
Goldstein et al.~\shortcite{goldstein2012video} enhances the feature track length with epipolar transfer.
The advantage of 3D and 2.5D methods is that they produce strong stabilization effects. 

2D methods generally perform efficiently and robustly via applying 2D linear transforms to video frames~\cite{gleicher2007re}.
Some early works~\cite{matsushita2006full,chen2008capturing} apply low-pass filters for motion smoothing while also providing full-frame capabilities via motion inpainting.
%Motion inpainting involves filling in missing pixels by referencing adjacent frames.
In more recent works, Grundmann et al.~\shortcite{grundmann2011auto} apply L1-norm optimization to compute smooth camera paths consisting of cinematography motions. 
They extended this idea to compensate for rolling shutter effects~\cite{grundmann2012calibration}.
Liu et al.~\shortcite{liu2013bundled} model the camera trajectory on multiple local camera paths for spatially variant motions, while they next proposed to smooth pixel profiles rather than feature tracks~\cite{liu2014steadyflow}.
This was extended to a more efficient smoothing of mesh profiles~\cite{liu2016meshflow}, and further extended to video coding~\cite{liu2017codingflow}.
Huang et al.~\shortcite{huang2018encoding} also use video coding to improve the efficiency of camera motion search.
Among other techniques, spatiotemporal optimization~\cite{wang2013spatially}, and user interactions~\cite{bai2014user} have been applied to generate smooth camera motion.
Online video stabilization methods~\cite{liu2016meshflow,wang2018deep,xu2018deep,liu2017codingflow} use historical frames generated from the model to estimate current stabilization parameters.
Although offline methods generally show better results, deep learning-based approaches~\cite{wang2018deep,xu2018deep} show promising results.

Recently we have seen the success of deep learning algorithms applied to various computer vision tasks.
It has been shown that CNNs in particular provide a promising direction in the video stabilization task as well.
Instead of explicitly computing a camera path, these methods model a supervised learning approach.
Wang et al.~\shortcite{wang2018deep} propose a supervised learning framework for video stabilization by defining the stability and temporal loss terms. 
This work also provides a labeled dataset consisting of unstable and stable video sets.
Xu et al.~\shortcite{xu2018deep} propose an adversarial network also trained in a supervised manner, where warping parameters are estimated by the adversarial network to generate stabilized frames.
Our method inherits the merits of 2D methods and performs video stabilization in the context of deep frame interpolation.
Due to this, our results involve no cropping, while minimal distortion is introduced.
Furthermore, our method is trained in an unsupervised manner, not requiring ground truth videos.

%%%%%%%%%%%%%%%%%%%%%%%%%%%%%%%%%%%%%%%%%%%%%%%%%%%%%%%%%%%%%%%%%%%%%
\section{Proposed Method: DIFRINT}
\label{method}
The key idea of our approach is to sequentially synthesize a middle frame between two frames of a given video, leading to visual stability.
Our approach can be considered as a 2D method, where each consecutive frames undergo frame interpolation via utilizing neighboring frames.
Throughout this text, `interpolated frame' and `middle frame' are synonymous and are used interchangeably.

Our method allows applying interpolation with multiple iterations, producing stronger stabilization effects.
The simplest way to implement such method is to simply devise a deep frame interpolation architecture.
As shown in Fig.~\ref{framework}~(b), given two frames, they are warped `toward' each other \emph{halfway} (denoted as `$\times 0.5$') via bidirectional optical flow (PWC-Net~\cite{Sun_CVPR_2018}). 
More specifically, $f_{i-1}$ is warped to $f_{i+1}$ halfway, while $f_{i+1}$ is also warped to $f_{i-1}$ halfway, producing warped frames $f_w^-$ and $f_w^+$. 
Thus, $f_w^-$ and $f_w^+$ represent the halfway points each originating from $f_{i-1}$ and $f_{i+1}$ respectively.
Then, these halfway point frames $f_w^-$ and $f_w^+$ are fed to a convolutional neural network (CNN), namely a U-Net architecture~\cite{ronneberger2015u} to generate the middle frame $f_{int}$.
The U-Net module is able to learn how information at different scales should be combined, guiding local high-resolution predictions with global low-resolution information.
Note that warped images may contain holes or unseen regions, which is why we use both $f_w^-$ and $f_w^+$ to complement each other, which the CNN essentially learns.

\begin{figure}
	\includegraphics[width=1\linewidth,keepaspectratio]{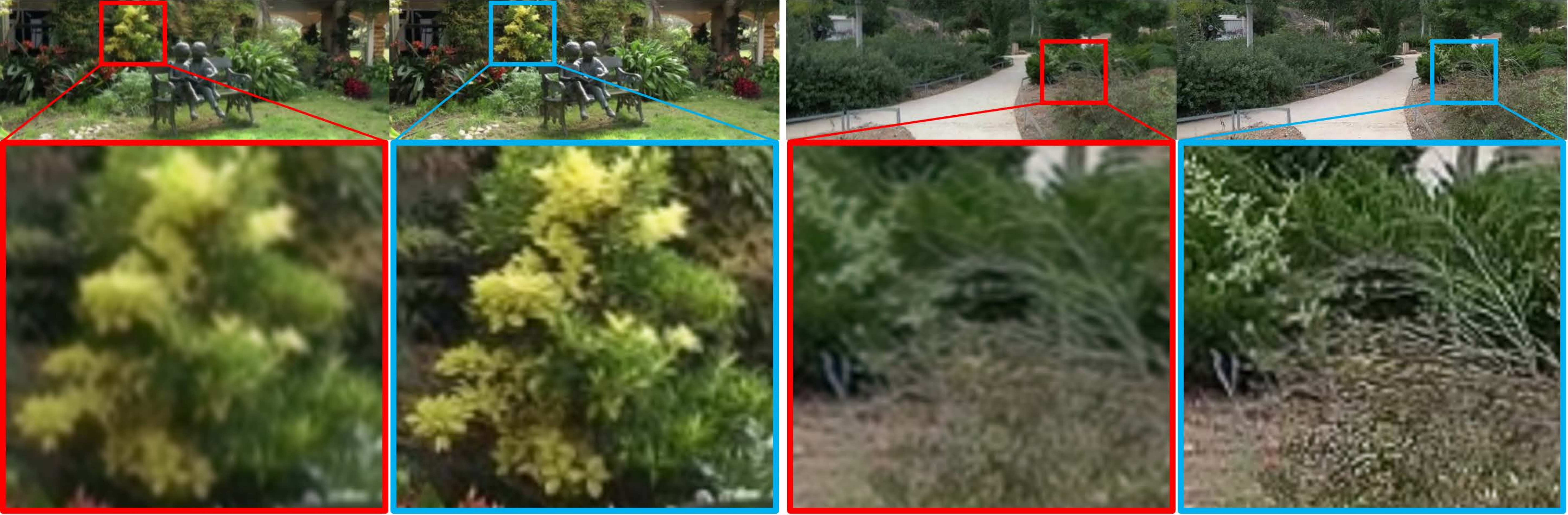}
	%\vspace{-5mm}
	\caption{Results of the interpolation only method (in red) and our full model (in blue).
	Magnified patch segments show that the full model preserves visual detail while the interpolation only method accumulates blur.}
	\label{blur}
	%\vspace{-5mm}
\end{figure}

%\subsection{Preventing blur accumulation}
The deep frame interpolation only method described above is prone to blur accumulation during multiple iterations as shown in Fig.~\ref{blur}.
This is due to the information loss of fine details through multiple iterations.
To prevent blur accumulation, we utilize the given original frame $f_i$ which is warped toward $f_{int}$, and fed along with $f_{int}$ to another CNN (ResNet architecture~\cite{he2016deep}) to produce the final middle frame $\hat{f}_i$.
The original frame $f_i$ is utilized for \emph{every} iteration.
Thus, even with multiple iterations, resampling error does not occur since the original frame contains such fine details.
The ResNet architecture is well suited for minimizing error at fine details through its residual learning mechanism.

As summary, frame interpolation occurs by warping two frames toward each other halfway, which are fed to a CNN, producing an `intermediate frame' $f_{int}$.
Then, $f_{int}$ is used as \emph{reference} to which the original frame $f_i$ is warped, and fed along with $f_{int}$ to another CNN, producing the final interpolated frame $\hat{f}_i$.
In the following subsections, we explain details on the training and testing schemes including loss functions and detailed settings.

\subsection{Unsupervised learning framework}
A major benefit of our approach is that it takes advantage of deep learning and frame interpolation to address video stabilization via unsupervised learning.
In the following, we explain how our model is trained and how it is applied during testing for stabilization.

\subsubsection{\textbf{Training scheme}}
The goal of our proposed framework is to generate middle frames without error accumulation.
Therefore, the goal of the training scheme is to properly train the framework such that it produces such interpolation quality.
To understand the training scheme, the testing scheme must be understood first.
As explained above, the actual interpolation is implemented via warping two frames halfway toward each other, and feeding them through a U-Net architecture, generating $f_{int}$.
To prevent any error or blurring, the original frame $f_i$ is utilized by warping it toward $f_{int}$, and feeding them altogether through a ResNet architecture, producing $\hat{f}_i$.
A problem arises when attempting to train this model: the ground truth middle frame does not exist.
We cannot simply use $f_i$ as ground truth since it is not guaranteed to be the halfway point between adjacent frames $f_{i-1}$ and $f_{i+1}$.
Thus, $\hat{f}_i$ does not have a ground truth to which we can compare and compute the loss for.

\begin{figure}
	\centering
	\subfloat[Training scheme]{{\includegraphics[width=1\linewidth,keepaspectratio]{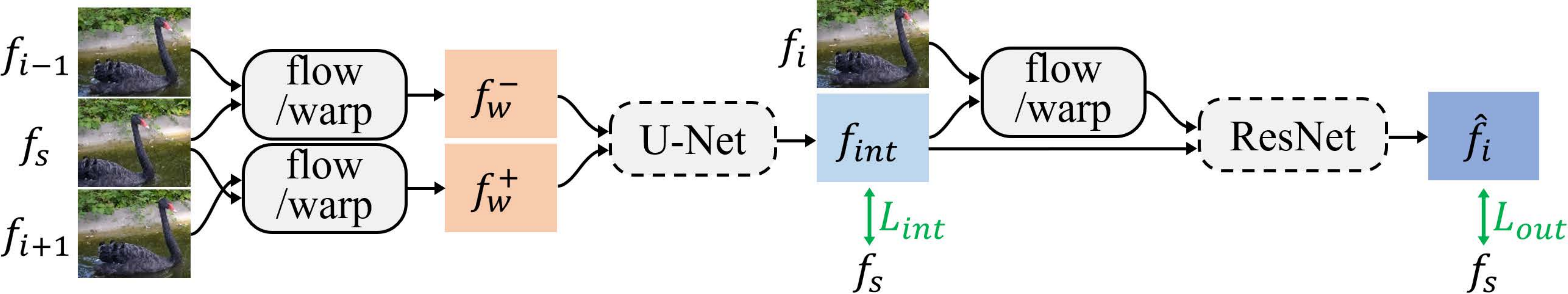}}}
	\vspace{-2mm}
	\subfloat[Testing scheme]{{\includegraphics[width=1\linewidth,keepaspectratio]{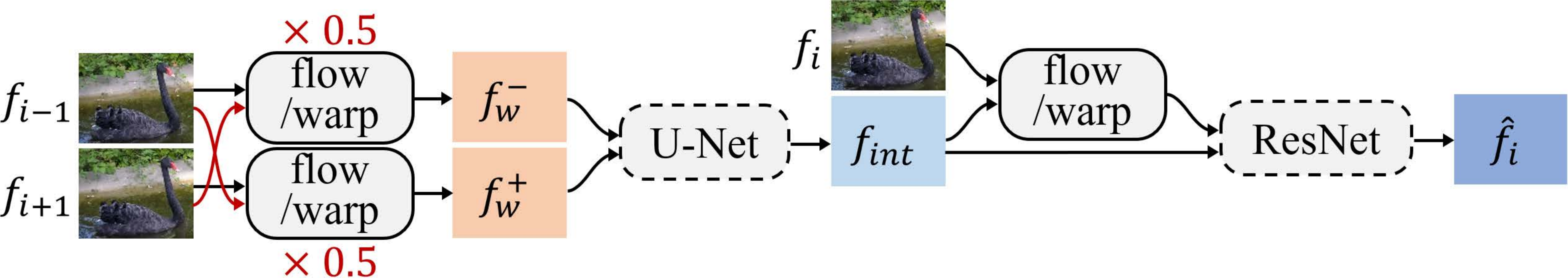}}}
	\caption{Our DIFRINT framework during (a) training and (b) testing. 
	The differences between training and testing is indicated in red. 
	Trainable components are expressed with dotted boundaries while fixed components are solid.
	The loss applied is shown in green.}
	\label{framework}
	%\vspace{-5mm}
\end{figure}

At this point, we define a pseudo-ground truth frame $f_s$ which is a spatially translated version of the original frame $f_i$.
Translation is done by a small random scale (at most one eighth of the frame width) in a random direction.
Training is done by warping the adjacent frames $f_{i-1}$ and $f_{i+1}$ toward $f_s$, and aiming to \emph{reconstruct} $f_s$.
In this way, the U-Net learns to reconstruct $f_s$, given two warped frames $f_w^-$ and $f_w^+$, while the ResNet utilizes the (warped) original frame $f_i$ to do the same.
This training scheme generalizes well to the testing scheme, which can be seen as reconstructing the \emph{virtual middle frame}.
An overview of the training scheme is shown in Fig.~\ref{framework}~(a).

\subsubsection{\textbf{Testing scheme}}
Once the training scheme properly trains the model through reconstructing the pseudo-ground truth $f_s$, the actual frame interpolation can be applied during the testing scheme.
The adjacent frames $f_{i-1}$ and $f_{i+1}$ are warped \emph{halfway} (by a factor of 0.5) toward each other as shown in Fig.~\ref{framework}~(b).
The technique of warping consecutive frames halfway or learning to predict the middle frame has been used in frame interpolation works~\cite{niklaus2018context,niklaus2017video,niklaus2017adaptive,meyer2015phase}.
In practice, given a video sequence, all consecutive frame triplets ($f_{i-1}$, $f_i$ and $f_{i+1}$) are used as input to our framework to generate stabilized frame outputs, while the first and last frames are left as they are.
We provide the option to further stabilize the generated video frames by iteratively applying our method on already interpolated frames.

\begin{figure}
	\includegraphics[width=1\linewidth,keepaspectratio]{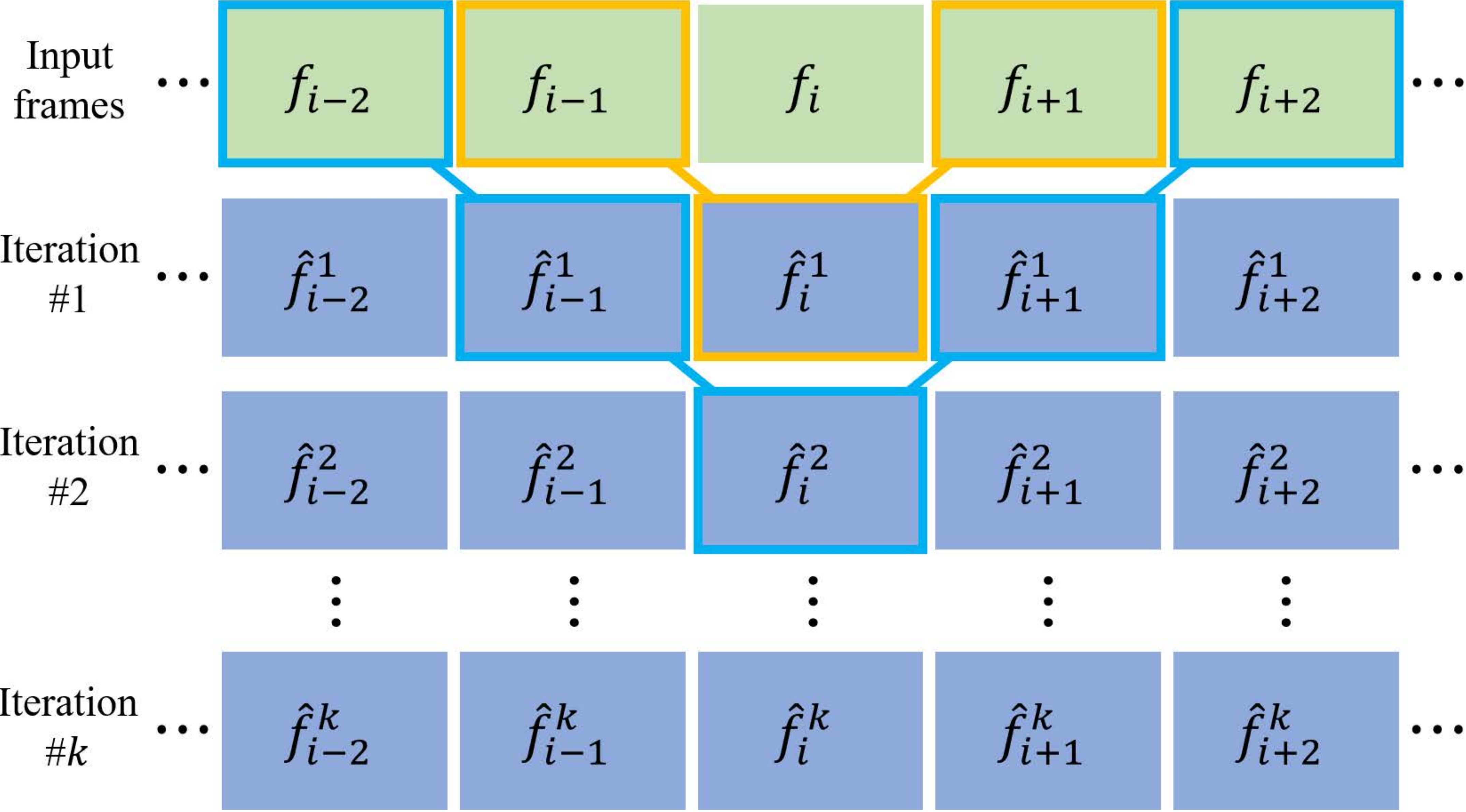}
	%\vspace{-5mm}
	\caption{Further iterative frame interpolation leads to stronger stabilization, since farther away frames affect the current interpolation (shown in blue connections).
	This leads to a more global stabilization than frames interpolated with lesser iterations (shown in yellow connections).}
	\label{iter}
	%\vspace{-5mm}
\end{figure}

\begin{figure}
	\includegraphics[width=1\linewidth,keepaspectratio]{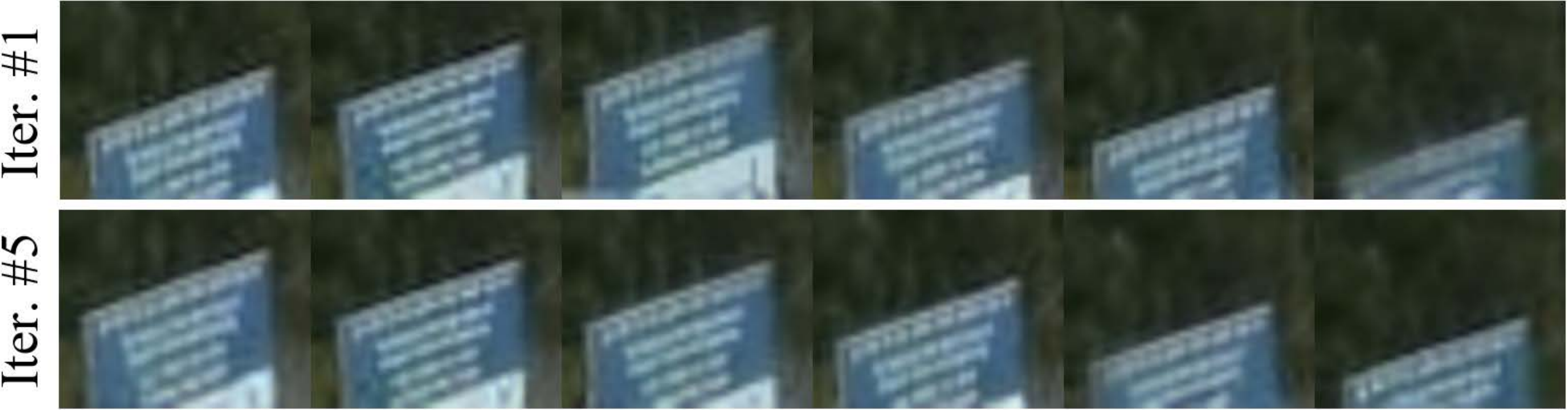}
	%\vspace{-5mm}
	\caption{Top: shows a stabilized sequence of a frame segment when one iteration is applied. Bottom: shows the results with five iterations. Results with five iterations show a more global stabilization whereas a single iteration shows temporally local fluctuation.}
	\label{iterqual}
	%\vspace{-5mm}
\end{figure}

Iterative frame interpolation produces stronger visual stability.
As shown in Fig.~\ref{iter}, since $\hat{f}^2_i$ has undergone two iterations, its spatial orientation is affected by as far as $f_{i-2}$ and $f_{i+2}$, whereas a single iteration producing $\hat{f}^1_i$ is only affected by $f_{i-1}$ and $f_{i+1}$.
As such, more iterations lead to a more global stabilization as shown in Fig.~\ref{iterqual}.
Here, we provide yet another parameter for adjusting stability, namely the skip parameter that modifies which frames to use for interpolation.
For example, the default interpolation uses $f_{i-1}$ and $f_{i+1}$ as adjacent frames (skip = 1), while setting the skip parameter to 2 leads to using $f_{i-2}$ and $f_{i+2}$ as input, and so on.
For triplets that do not have skipped adjacent frames (e.g. close to the first or last frames) are assigned smaller skip parameters.

\subsubsection{\textbf{Loss functions}}
To train our network components, we use the pixel-wise color-based loss function.
Since $\ell^2$-loss has been reported to produce blurry results~\cite{goroshin2015learning,mathieu2016deep,long2016learning}, we employ an $\ell^1$-loss function defined as follows:
\begin{equation}
\label{loss1}
L_1=\left\| f_s - \hat{f}_i \right\|_1
\end{equation}
where $L_1$ is the loss between the pseudo-ground truth frame $f_s$ and the output frame $\hat{f}_i$.
We also consider the perceptual loss function utilizing the response from the \texttt{relu4\_3} layer of VGG-19~\cite{simonyan2014very}:
\begin{equation}
\label{loss2p}
L_p=\left\| \phi{(f_s)} - \phi{(\hat{f}_i)} \right\|^2_2
\end{equation}
where $\phi$ denotes the feature vector.
We take the sum of $L_1$ and $L_p$ as the final loss:
\begin{equation}
\label{lossOut}
L_{out}=L_1+L_p
\end{equation}
Although the final loss $L_{out}$ is enough to train the whole network, we have found that 
applying the same loss to $f_{int}$, speeds up training and leads to better performance:
\begin{equation}
\label{lossInt}
L_{int}=\left\| f_s - f_{int} \right\|_1 + \left\| \phi{(f_s)} - \phi{(f_{int})} \right\|^2_2
\end{equation}
Since $f_{int}$ also essentially aims to reconstruct $f_s$, it is safe to apply this loss, which in effect explicitly trains the U-Net component.

\begin{figure}
	\includegraphics[width=1\linewidth,keepaspectratio]{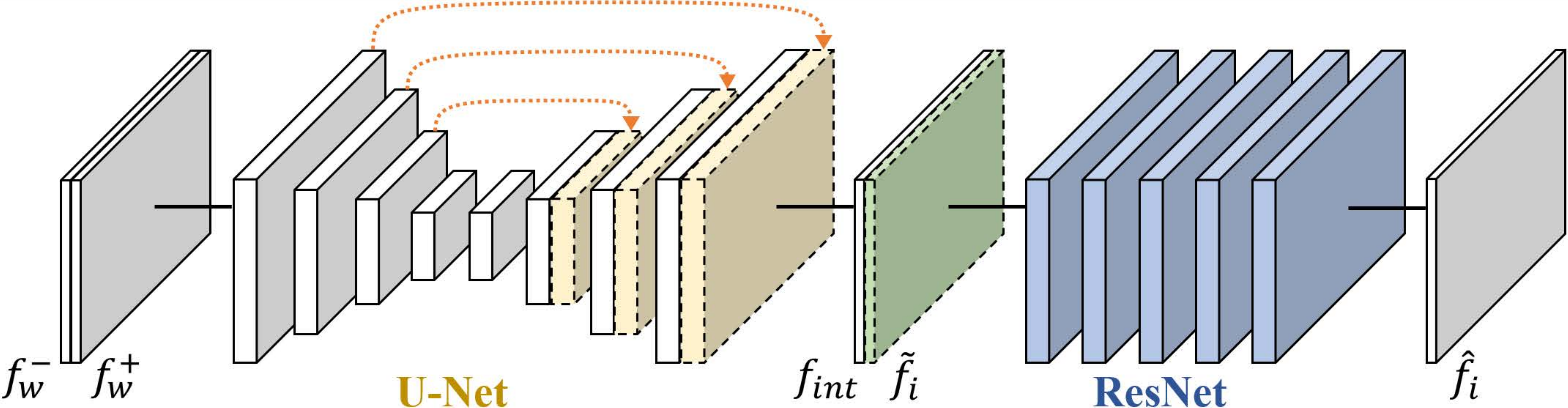}
	%\vspace{-5mm}
	\caption{The U-Net and ResNet architectures and the feed forward process are shown.
	The adjacent warped frames $f_w^-$ and $f_w^+$ are concatenated and fed through the U-Net, producing $f_{int}$.
	The original frame $f_i$ is warped toward $f_{int}$, denoted as $\tilde{f}_i$, and concatenated.
	It is then fed through the ResNet, producing $\hat{f}_i$.}
	\label{network}
	%\vspace{-5mm}
\end{figure}

\begin{figure*}
	\includegraphics[width=1\linewidth,keepaspectratio]{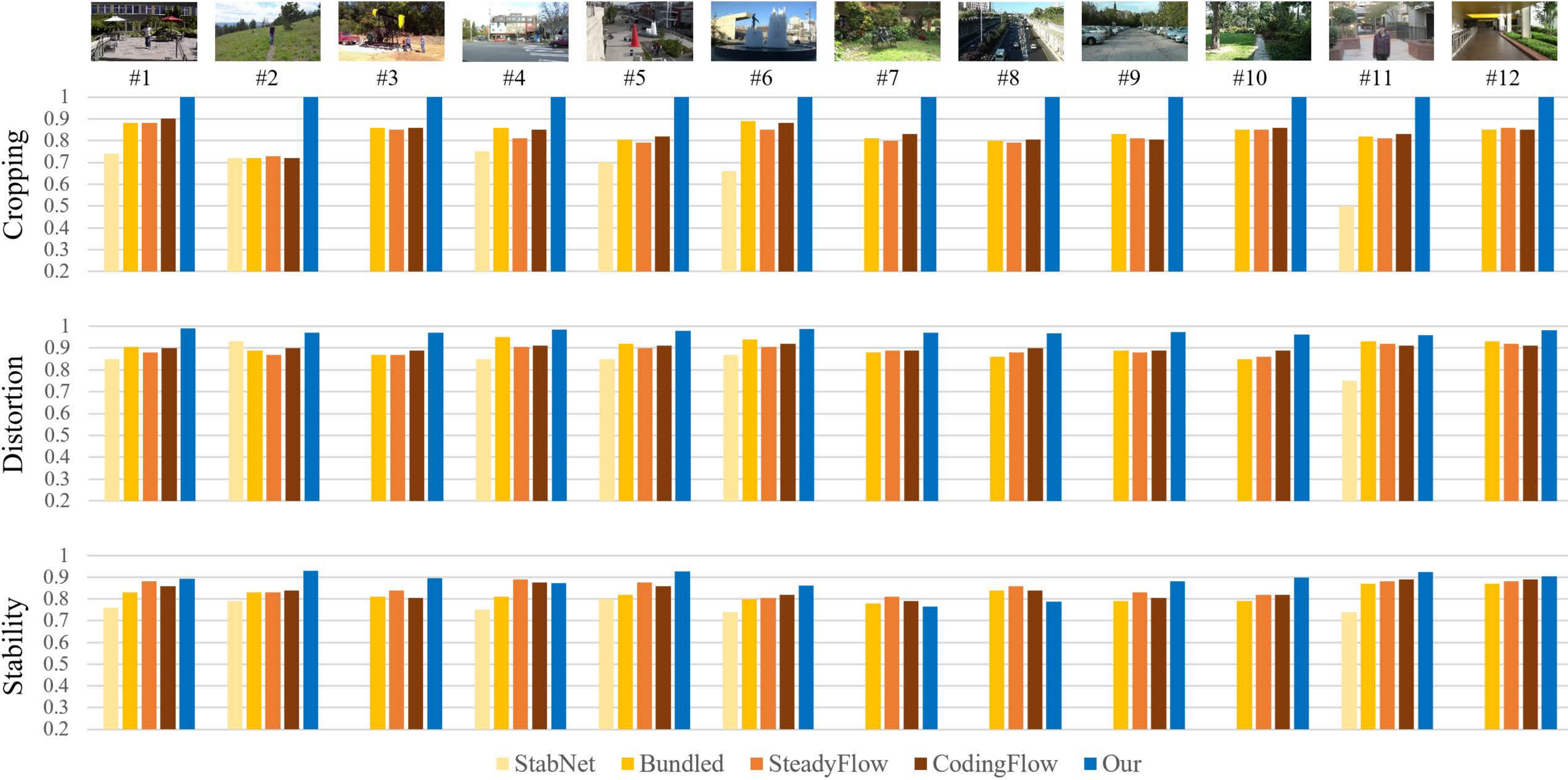}
	\vspace{-5mm}
	\caption{Quantitative comparisons with the state-of-the-art methods on publicly available video data.
	Comparisons are done against \cite{wang2018deep}, \cite{liu2013bundled}, \cite{liu2014steadyflow}, \cite{liu2017codingflow} from left to right.}
	\label{quant}
	\vspace{-2mm}
\end{figure*}

\subsection{Implementation}
Our framework is implemented via PyTorch.
%We implemented the warping layer and random-shift layer using CUDA and the grid sampler from cuDNN.
The training process using two NVIDIA Titan X (Maxwell) takes about 1 day, while generating a $1280\times720$ frame takes 0.07 seconds, yielding near real-time of 15 fps.
The generation process includes all three optical flow estimations, three warping layers, and feed forwarding through our U-Net and ResNet architectures.
In practice, making use of the multi-stream capability of CUDA, enables paralleled generation of frames as long as the previous iteration has done generating the two frames for input.

Within our proposed framework, the optical flow estimator (PWC-Net) is fixed (not trainable), while the trainable networks U-Net and ResNet are designed to produce exceptional quality with high speed.
As shown in Fig.~\ref{network}, the U-Net architecture employs three skip connections (dotted arrows) among scaled features.
The architecture contains $3\times3$ convolutional layers, while hidden feature channels are of size 32, down/upscaled by a factor of 2.
The U-Net produces the intermediate frame $f_{int}$, and the original frame $f_i$ is warped toward it, denoted as $\tilde{f}_i$ (shown in green).
These are concatenated and fed through the ResNet architecture with five residual blocks (shown in blue).
The residual blocks employ $1\times1$ convolutional layers (channel size 32) to minimize any noise from adjacent pixels during reconstruction of the final output $\hat{f}_i$.
The $1\times1$ kernels also speed up the feed forward process.
In addition, all convolutional layers of our entire trainable components employ gated convolutions~\cite{yu2018free} ($3\times3$) which have shown superior quality in image inpainting tasks. 
Gated convolutions allow dynamic feature selection for each spacial (pixel) location and each channel.
The gated convolutions are well suited to our problem where input images may contain holes or unseen regions after warping.

\subsubsection{\textbf{Training settings}}
We train our entire framework on frames of size $256\times256$ using the Adam optimizer with $\beta_1=0.9$ and $\beta_2=0.999$, a learning rate of 0.001, and mini-batch size of 16 samples.
We use the DAVIS dataset~\cite{Perazzi2016} since it contains diverse and dynamic scenes, and since our approach does not need any ground truth video sets.
We utilize approximately 8,000 training examples, and train the architecture for 200 epochs, with a linear decay applied starting from epoch 100.
To eliminate potential dataset bias, we also augment the training data on the fly by choosing random patches of size $256\times256$ from the full original frames.
We also randomly apply horizontal and/or vertical flips to each patch. 

\subsubsection{\textbf{Testing settings}}
During testing, the full resolution frames are fed into our framework to generate stabilized frame sequences.
We observe that our framework generalizes well to larger full resolution images.
As default, we set the number of iterations to 5 and skip parameter to 2 throughout all of our results.

%%%%%%%%%%%%%%%%%%%%%%%%%%%%%%%%%%%%%%%%%%%%%%%%%%%%%%%%%%%%%%%%%%%%%
\section{Experiments}
To thoroughly evaluate our method, we conduct an extensive comparison to representative state-of-the-art stabilization methods.
To the best of our knowledge, compared to previous works, our evaluation has the largest scale in terms of the number of comparison baselines and the number of test videos.
We also provide visual comparisons and ablation study visualizations.
% comprised of both quantitative and qualitative comparisons. -> quantitative ablation may show high score for poor results, may raise question.
Finally, we conduct a user preference study against commercial methods.
Full video results can be found in our supplementary video, which contains results of $640\times360$, $1280\times720$, and $1920\times1080$ resolutions.

\begin{figure*}
	\includegraphics[width=1\linewidth,keepaspectratio]{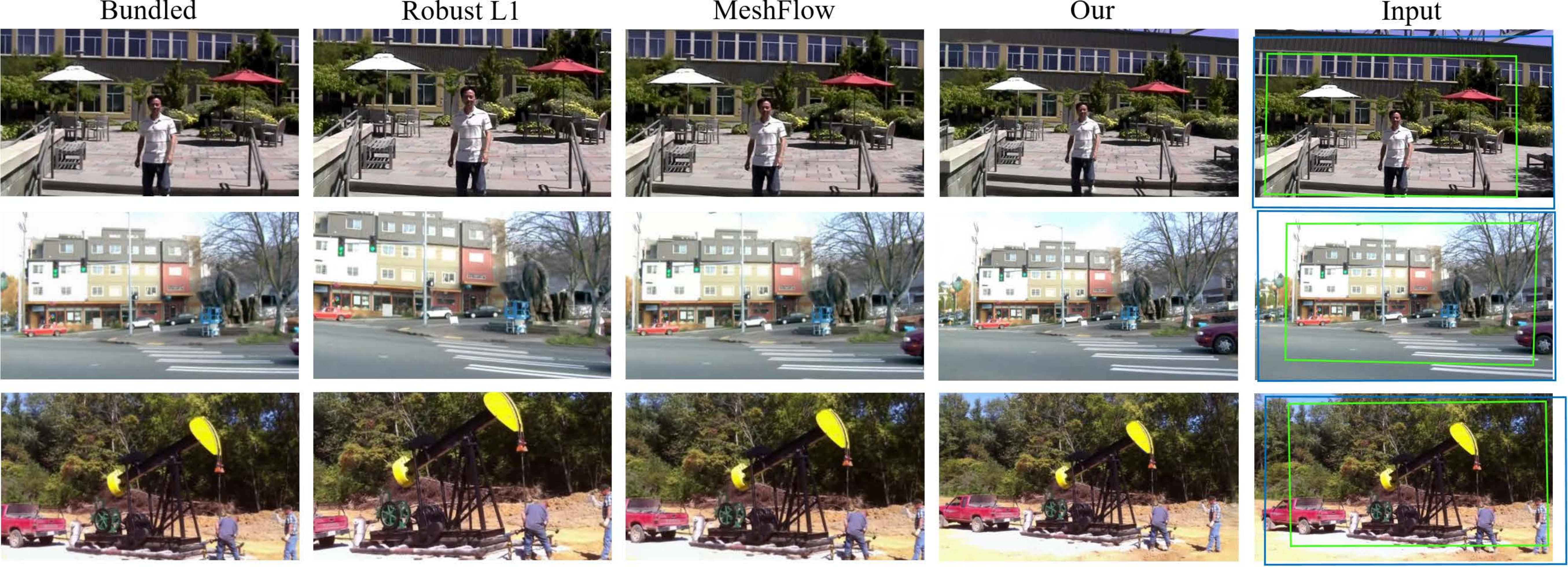}
	%\vspace{-5mm}
	\caption{Examples of stabilized frames from \cite{liu2013bundled}, \cite{grundmann2011auto}, \cite{liu2016meshflow}, our method, and the corresponding input frame (from left to right respectively). The green boxes show the cropped region of the largest among the three comparisons. Our method does not show any cropping; only stabilized ``shifts,'' generating image regions which are unseen from the input frame, shown as blue boxes.}
	\label{crop}
	\vspace{-2mm}
\end{figure*}

\subsection{Quantitative evaluation}
Video stabilization methods are typically assessed by three metrics: \emph{cropping ratio}, \emph{distortion value}, and \emph{stability score}.
For each of these metrics, a good result is closer to the value of 1.0.
We briefly explain each metric as follows.

\textbf{Cropping ratio} measures the remaining image area after cropping away missing view (black) boundaries.
Larger ratio signifies less cropping and thus better video quality.
A homography is fit between the input and output for each frame, and the scale component is extracted.
The average scale across the entire video yields the cropping ratio metric.

\textbf{Distortion value} gives the anisotropic scaling of the homography between an input and output frame.
It can be computed by the ratio of the two largest eigenvalues of the affine part of the homography.
Among homographies for all frames, the worst ratio is chosen for the distortion value metric.

\textbf{Stability score} measures the overall smoothness of the output video.
For this metric, a frequency domain analysis is used.
Similar to \cite{liu2013bundled}, the camera path is used to compute the value of stability.
%First, a camera path is defined by the accumulated homographies between successive frames: $P^t=H_0H_1 \dots H_{t-1}$ where $H_i$ denotes the homography between frames $f_i$ and $f_{i+1}$ of the output video.
%First, the vertex profiles are extracted from the stabilized video.
%Then, each vertex profile is analyzed in the frequency domain. 
Two 1D profile signals can be made from extracting the translation and rotation components.
The ratio of the sum of lowest (2nd to 6th) frequency energies and the total energy is computed, and the final stability score is obtained by taking the minimum.

\subsubsection{Quantitative results}
In total, we compare 13 baselines including state-of-the-art methods and commercial algorithms.
We used a total number of 25 publicly available videos~\cite{liu2013bundled} for evaluation.
Due to space limitation, we show 12 representative results of the top performing baselines (including one deep online method and three 2D methods) in comparison to our results in Fig.~\ref{quant}.
The representative results were chosen by following \cite{liu2017codingflow}, since it contains the most number of comparisons with top performing baselines.
We present our results accordingly.
For the entire comparison to 13 baselines using 25 videos, please refer to the supplementary material.
For actual video comparisons, please refer to our supplementary video.
We implemented the code for the aforementioned metrics and to check its validity, we have undergone sanity checks on several published video results and consistently achieved reported scores.

One can observe that our method shows the best performances for the majority of the 12 videos shown.
In particular, our method shows cropping ratios of 1.0 and maintains distortion values close to 1.0 for all videos.
Our method shows the highest stability scores for the majority of the videos, and a few comparable results.

\begin{figure}
	\includegraphics[width=1\linewidth,keepaspectratio]{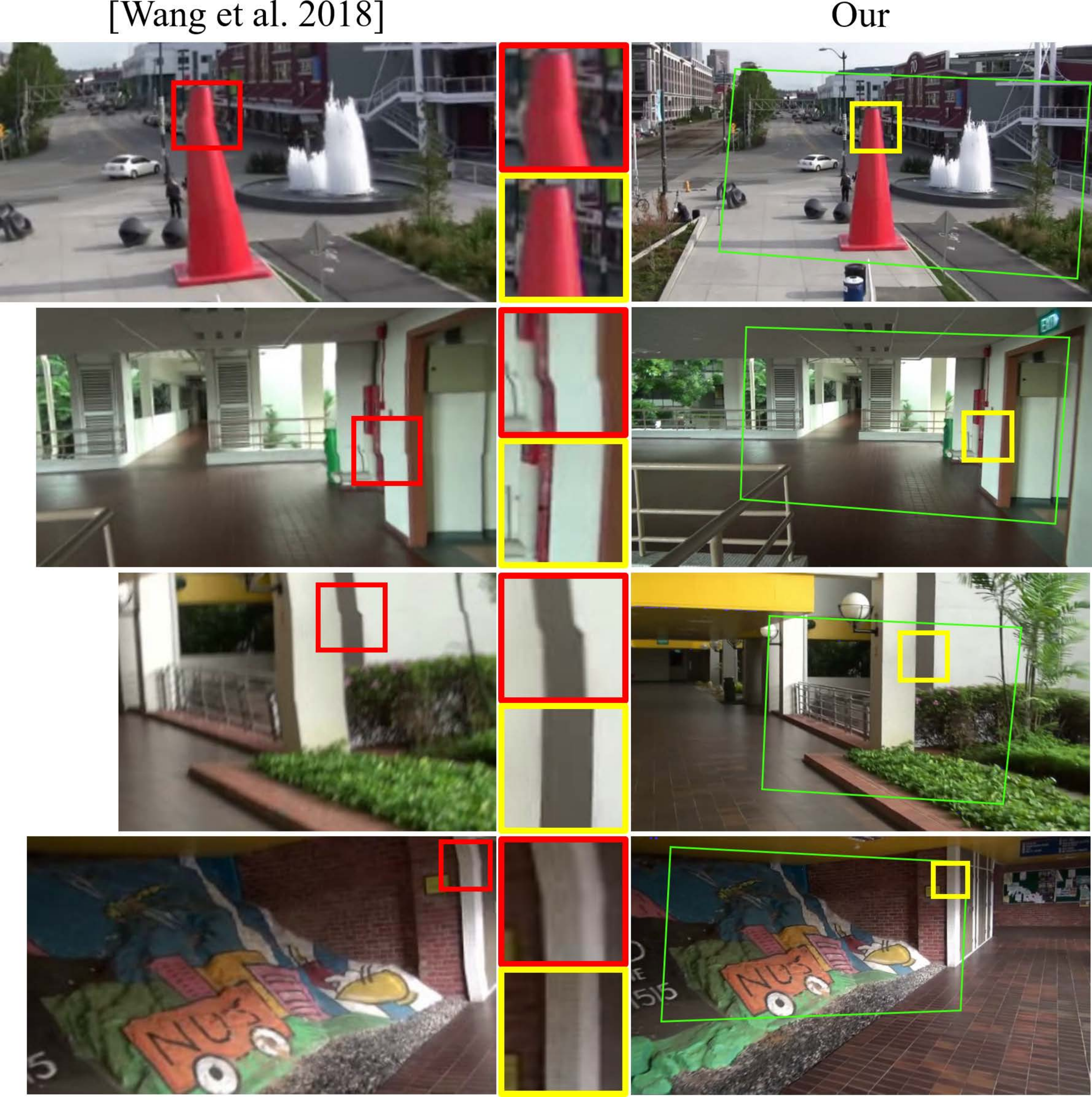}
	%\vspace{-5mm}
	\caption{Comparison to StabNet~\cite{wang2018deep} results. The red and yellow boxes are magnified and displayed to visually compare distortion artifacts. The green boxes show the cropped regions of ~\cite{wang2018deep} with respect to ours.}
	\label{distortion}
	\vspace{-5mm}
\end{figure}

\subsection{Visual comparison}
\begin{figure}
	\includegraphics[width=1\linewidth,keepaspectratio]{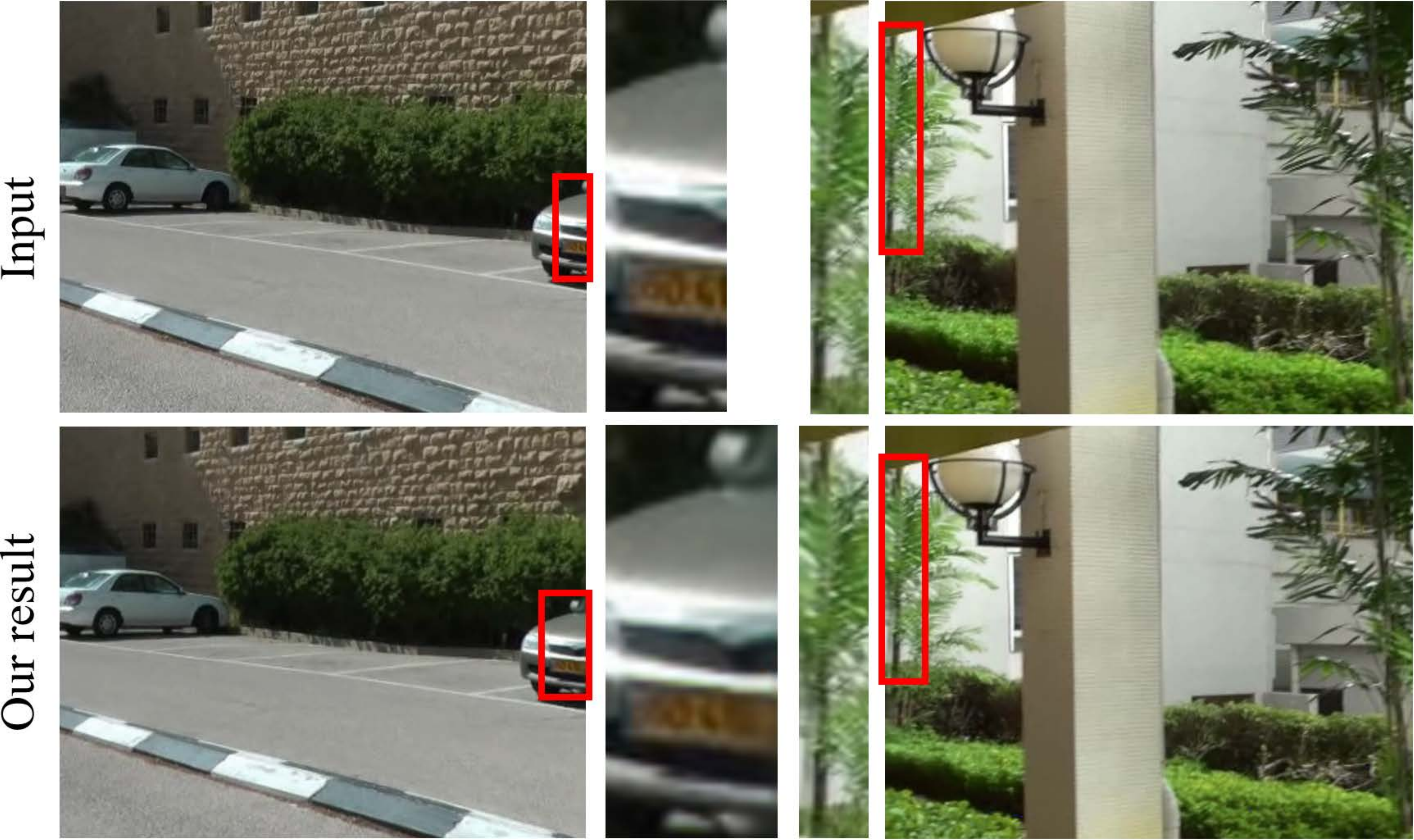}
	\vspace{-5mm}
	\caption{As a result of deep iterative interpolation, our method can generate unseen regions (i.e. missing view) at image boundaries.}
	\label{unseen}
	\vspace{-3mm}
\end{figure}

For qualitative evaluation, we visually compare our results with several state-of-the-art methods.
The state-of-the-art results were obtained from videos published by the authors or via available code online. 
Fig.~\ref{crop} shows examples of stabilized frames from three videos using \cite{liu2013bundled}, \cite{grundmann2011auto}, \cite{liu2016meshflow}, and our method.
Compared to the corresponding input frames, all three baselines have some degree of cropping and enlargement effect, whereas our method does not induce any cropping nor zoom-in effects.
Our results seem as though the camera has shifted to stabilize the video frame, generating unseen regions rather than cropping them out.
This generation of unseen regions can be seen in more detail in Fig.~\ref{unseen}.
As a result of deep iterative frame interpolation, our results can generate content at image boundaries which cannot be seen from the corresponding input frame.

\begin{figure*}
	\includegraphics[width=1\linewidth,keepaspectratio]{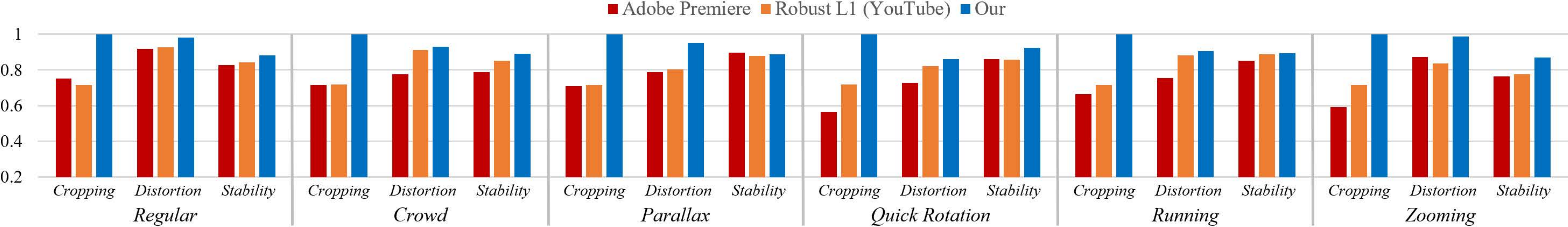}
	\vspace{-5mm}
	\caption{Quantitative comparisons with Adobe Premiere Pro CC 2017, \cite{grundmann2011auto}, and our method on 6 video categories.}
	\label{us_quant}
	%\vspace{-5mm}
\end{figure*}

\begin{figure*}
	\includegraphics[width=1\linewidth,keepaspectratio]{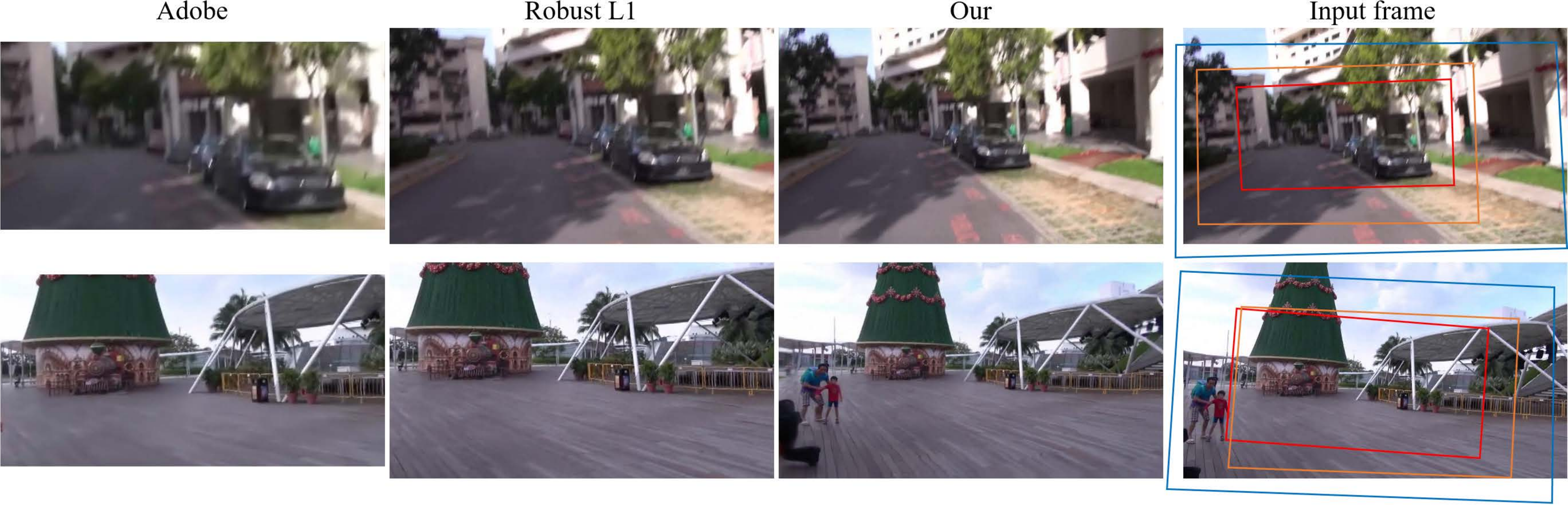}
	\vspace{-7mm}
	\caption{Visual comparisons to Adobe and Robust L1 results on challenging camera motion segments (from the \emph{Running} video category).}
	\label{us_qual}
	%\vspace{-5mm}
\end{figure*}

Since our method uses frame interpolation techniques for stabilization, global image distortion is significantly low.
We compare our method to the state-of-the-art online video stabilization method \cite{wang2018deep} in Fig.~\ref{distortion}.
We observe that our method introduces lower levels of distortion and contains more of the original content.

\begin{figure}
	\includegraphics[width=1\linewidth,keepaspectratio]{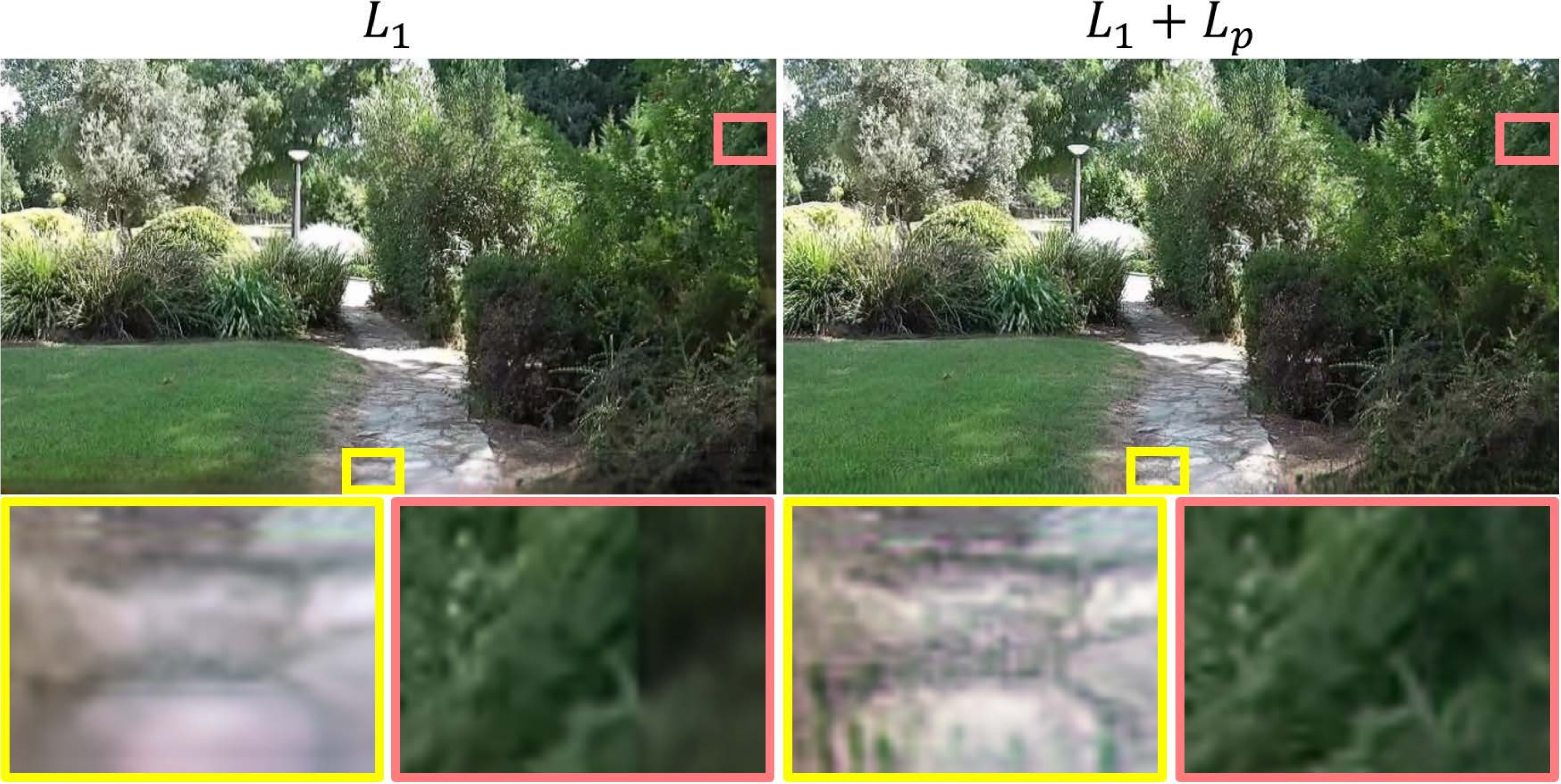}
	\vspace{-5mm}
	\caption{Ablation study comparing the visual effects on using just the $\ell^1$-loss, and using an additional perceptual loss function.
	Only using the $\ell^1$-loss produces blurry and darker artifacts near the boundaries.}
	\label{ablation}
	\vspace{-3mm}
\end{figure}

As an ablation study, we compared the effect of using the perceptual loss to only using the $\ell^1$-loss function.
Using the perceptual loss showed marginal improvements in terms of the stabilization metrics (cropping ratio, distortion value, and stability score).
In contrast, using the additional perceptual loss showed improvements in visual quality as shown in Fig.~\ref{ablation}.
Due to iterative reconstruction, only using the $\ell^1$-loss leads to blurry and darker artifacts near the~boundaries.
Additionally, we conduct quantitative ablation studies on varying number of iterations and skip parameter, shown in the supplementary material.

%To show visual comparisons, we display examples of the mean and standard deviation of 5 consecutive video frames in Fig. [].
%Generally, as opposed to our method, other baselines show relatively blurred mean frames and larger standard deviations.

\subsection{User study}
To evaluate against commercial algorithms, we conduct a user study comparing our method to the warp stabilizer in Adobe Premiere Pro CC 2017.
Since YouTube no longer supports the stabilization option, we cannot directly compare our method to the YouTube's algorithm. 
However, to the best of our knowledge, the YouTube stabilizer is based on \cite{grundmann2011auto}.
Thus, we additionally compare our results to this method.

\begin{figure}
	\includegraphics[width=1\linewidth,keepaspectratio]{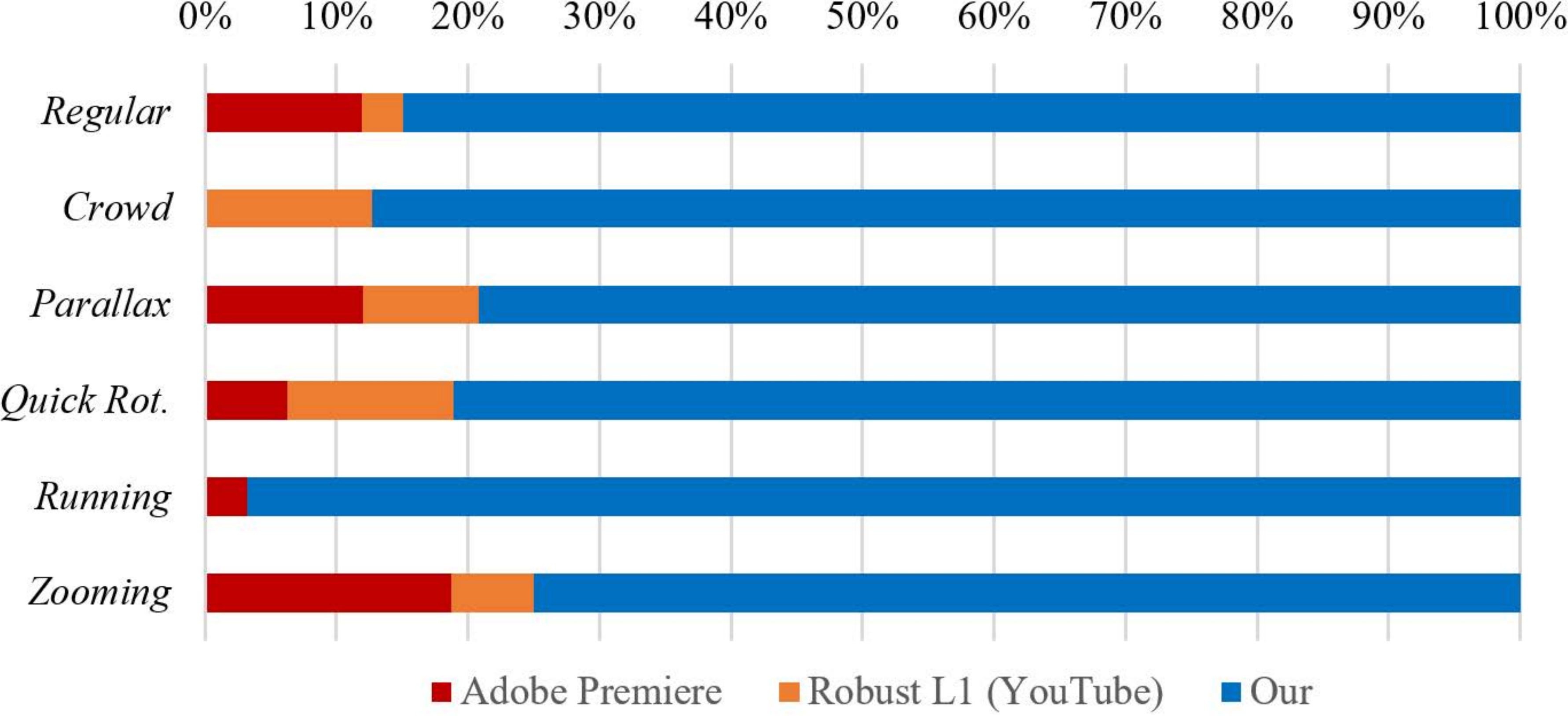}
	\vspace{-5mm}
	\caption{User preference test results on 6 video categories.}
	\label{us}
	\vspace{-3mm}
\end{figure}

First we conduct a category-wise comparison as conducted in \cite{xu2018deep}, with 6 categories: \emph{Regular}, \emph{Crowd}, \emph{Parallax}, \emph{Quick Rotation}, \emph{Running}, and \emph{Zooming} as shown in Fig.~\ref{us_quant}.
The 6 categories contain challenging scenarios which include many moving people, large difference in parallax, quick camera rotation, severe shaking due to running, and shaking due to zooming.
For all of these challenging scenarios, similar to the quantitative comparison to state-of-the-art algorithms, our method shows overall better performance than the commercial algorithms. 
Especially for challenging scenes with severe shaking, such as from the \emph{Running} category, the commercial algorithms show extensive cropping of the margins to compensate for successive missing view.
In contrast, as shown in Fig.~\ref{us_qual}, our method retains important content.
Moreover, we noticed that the commercial methods introduce slight wobbling artifacts and distortions especially at video segments with severe shaking.
%For actual video results, please refer to our video supplement.

The user study was conducted with 42 participants as a preference test among results from the Adobe Premiere, Robust L1 (YouTube), and our method.
Every participant evaluated 18 sets of videos (3 videos per category) in a three-alternative forced choice manner, given the original input video as reference. 
Each participant was asked to select the best result among the 3 choices, which were displayed in random order.
The participants were unaware of how each stabilized video was generated. 
The videos could be played, replayed, and paused for thorough evaluation.
To eliminate any bias, we asked the participants to select the best video in terms of stability and overall visual quality, and did not suggest any additional evaluation guidelines.
The preference results are shown in Fig.~\ref{us}.
We can observe that our method has the majority preference for all 6 video categories.
For severely unstable videos from the \emph{Running} category, our method is especially preferred.
For videos from the \emph{Zooming} category, the undesirable zoom-in effect from the commercial methods may have been relatively unnoticed due to the already zoomed-in content.

%%%%%%%%%%%%%%%%%%%%%%%%%%%%%%%%%%%%%%%%%%%%%%%%%%%%%%%%%%%%%%%%%%%%%
\section{Limitations \& Discussion}
A minor advantage of our method is that the user can adjust the number of iterations and the skip parameter to their liking.
Some users may prefer to leave some degree of instability for certain types of videos.
Instead of a one-size-fits-all method applied to every video, this may provide some freedom of manipulation to the user.
However, our method applies linear interpolation to all motion types without discrimination.
A possible future direction would be to dynamically determine motion types of different video segments and apply stabilization accordingly.
A simple method would be to measure the amount of camera shake via inter-frame homographies and apply different number of stabilization iterations to each segment depending on the amount of instability.

\begin{figure}
	\includegraphics[width=1\linewidth,keepaspectratio]{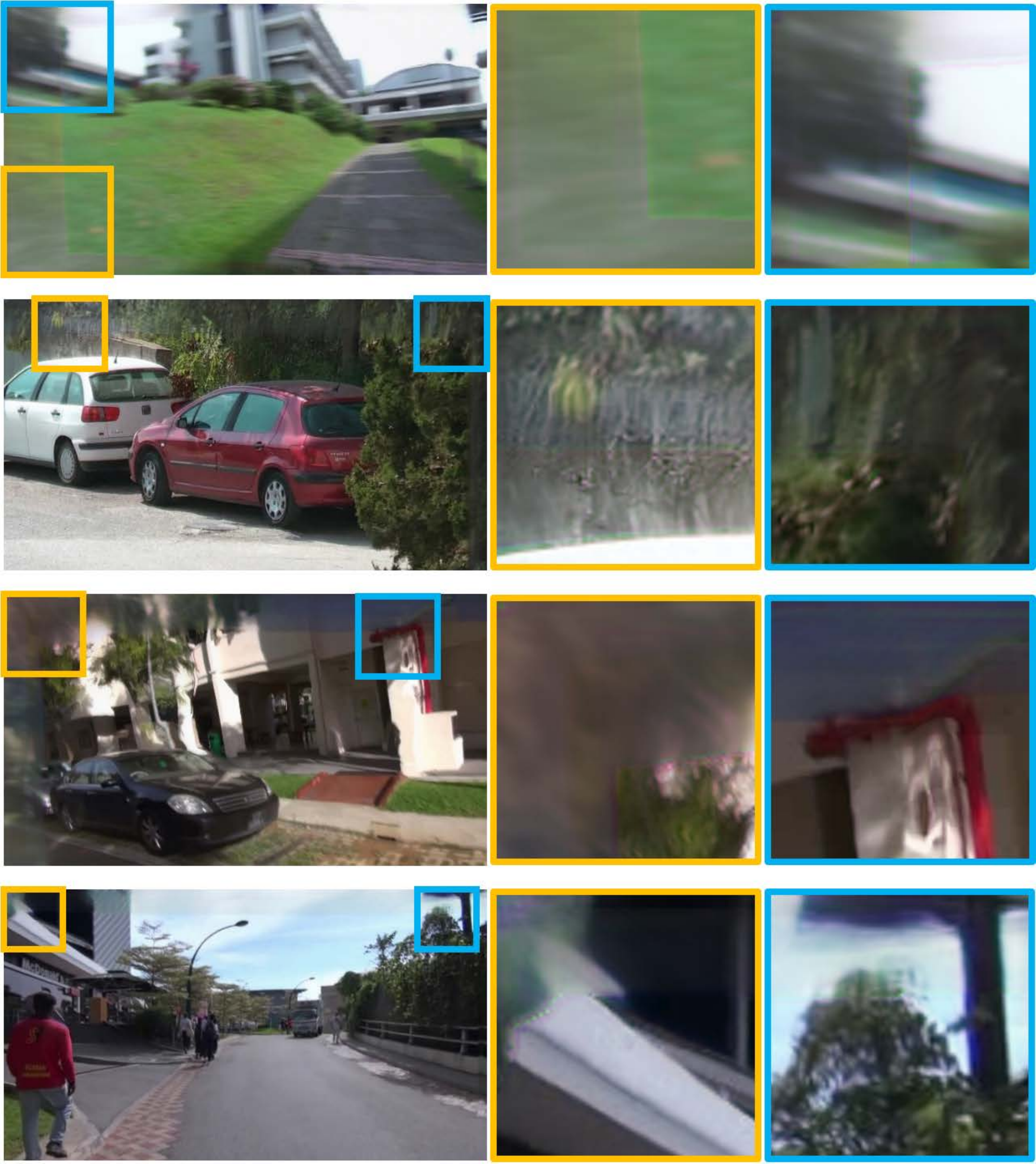}
	\vspace{-5mm}
	\caption{Failure cases occur on severe camera shakes, such as quick rotation, zoom-in and running scenes. 
	The boundary regions show severe blur due to unseen content.}
	\label{failurecase}
	\vspace{-2mm}
\end{figure}

Another limitation of our approach is that it may introduce blur at the image boundaries on severely shaking videos.
The majority of the frame content will be preserved since our method utilizes the original frames.
Since severe camera shake causes large unseen regions, previous methods tend to crop out large regions (while introducing wobbling artifacts), while our method shows blur artifacts toward the frame boundaries.
Although the user study suggests a preference to blur artifacts over severe cropping (\emph{Running} category of Fig.~\ref{us}), severe camera shakes remain as challenging scenarios.
Fig.~\ref{failurecase} shows examples of severe blur as failure cases for challenging videos.
A future direction would be to explicitly exploit multiple neighboring frames to reconstruct large unseen regions.
The challenge would be to achieve sufficient reconstruction quality while also handling memory issues.

%%%%%%%%%%%%%%%%%%%%%%%%%%%%%%%%%%%%%%%%%%%%%%%%%%%%%%%%%%%%%%%%%%%%%
\section{Conclusion}
When a user decides to apply video stabilization to one's personal video, the user must currently take into account some degree of cropping and enlargement, leading to loss of content and unwanted zoom-in effects.
We propose a method that aims to \emph{only} stabilize personal videos without any unwanted effects.
Furthermore, our lightweight framework enables near real-time computational speed.
Our method is an unsupervised deep learning approach to video stabilization via iterative frame interpolation, resulting in stabilized full-frame videos with low visual distortion.
From our proposed method, we hope iterative frame interpolation to be considered useful for the task of video stabilization.

%\textbf{Acknowledgments} This work was supported by the IITP grant funded by the Korea government (MSIT) (2017-0-01780).

\section*{Acknowledgments}
This work was supported by Institute for Information \& communications Technology Promotion (IITP) grant funded by the Korea government (MSIT) (2017-0-01780, The technology development for event recognition/relational reasoning and learning knowledge based system for video understanding).

% Bibliography
\bibliographystyle{ACM-Reference-Format}
\bibliography{sample-bibliography}

\end{document}